\def\year{2022}\relax
\documentclass[letterpaper]{article} 
\usepackage{aaai22}  
\usepackage{times}  
\usepackage{helvet}  
\usepackage{courier}  
\usepackage[hyphens]{url}  
\usepackage{graphicx} 
\usepackage{natbib}  
\usepackage{caption} 
\DeclareCaptionStyle{ruled}{labelfont=normalfont,labelsep=colon,strut=off} 
\usepackage{tikz}
\usepackage{algorithm}
\usepackage{algorithmic}

\usepackage{hyperref}

\usepackage{newfloat}
\usepackage{listings}
\floatstyle{ruled}
\newfloat{listing}{tb}{lst}{}
\floatname{listing}{Listing}
\nocopyright

\title{Robust Unstructured Knowledge Access in Conversational Dialogue with ASR Errors}
\author {
    Yik-Cheung Tam,\textsuperscript{\rm 1}
    Jiacheng Xu, \textsuperscript{\rm 1}
    Jiakai Zou \textsuperscript{\rm 2}
    Zecheng Wang \textsuperscript{\rm 1}
    Tinglong Liao \textsuperscript{\rm 1}
    Shuhan Yuan \textsuperscript{\rm 1}
}
\affiliations {
    \textsuperscript{\rm 1} New York University Shanghai\\
    \textsuperscript{\rm 2} New York University\\
    \{yt2267, jx1038, jz3714, zw2374, tl2564, sy2448\}@nyu.edu
}

\newcommand\copyrighttextieee{
  \footnotesize \textcopyright © 2022 IEEE.  Personal use of this material is permitted. Permission from IEEE must be obtained for all other uses, in any current or future media, including reprinting/republishing this material for advertising or promotional purposes, creating new collective works, for resale or redistribution to servers or lists, or reuse of any copyrighted component of this work in other works.}
\newcommand\copyrightnotice{
\begin{tikzpicture}[remember picture,overlay]
\node[anchor=south,yshift=10pt] at (current page.south) {\fbox{\parbox{\dimexpr\textwidth-\fboxsep-\fboxrule\relax}{\copyrighttextieee}}};
\end{tikzpicture}%
}

\begin{document}

\maketitle

\begin{abstract}
Performance of spoken language understanding (SLU) can be degraded with automatic speech recognition (ASR) errors. We propose a novel approach to improve SLU robustness by randomly corrupting clean training text with an ASR error simulator, followed by self-correcting the errors and minimizing the target classification loss in a joint manner. In the proposed error simulator, we leverage confusion networks generated from an ASR decoder without human transcriptions to generate a variety of error patterns for model training. We evaluate our approach on the DSTC10 challenge targeted for knowledge-grounded task-oriented conversational dialogues with ASR errors. Experimental results show the effectiveness of our proposed approach, boosting the knowledge-seeking turn detection (KTD) F1 significantly from 0.9433 to 0.9904. Knowledge cluster classification is boosted from 0.7924 to 0.9333 in Recall@1. After knowledge document re-ranking,
our approach shows significant improvement in all knowledge selection metrics, from 0.7358 to 0.7806 in Recall@1, from 0.8301 to 0.9333 in Recall@5, and from 0.7798 to 0.8460 in MRR@5 on the test set. In the recent DSTC10 evaluation, our approach demonstrates significant improvement in knowledge selection, boosting Recall@1 from 0.495 to 0.7144 compared to the official baseline. Our source code is released in GitHub~\footnote{\url{https://github.com/yctam/dstc10_track2_task2.git}}.
\end{abstract}
%
%
\section{Introduction}
\label{sec:intro}
Knowledge-grounded task-oriented dialogue modeling has drawn a lot of attention in recent years due to its practical applications. In addition to dialogue state tracking in conventional dialogue, users can ask questions about a referenced restaurant of a dialogue such as ``does the hotel offer free wifi?'', requiring accessing an external knowledge base for answers. Virtual personal assistants listen to users' speech input with disfluencies or barge-ins, giving additional challenge to robust language understanding. Although accuracy of automatic speech recognition (ASR) has been improved significantly, ASR is not perfect and generates mis-recognized tokens. According to the Dialogue System Technology Challenge (DSTC10~\cite{kimdstc10}), a deep neural model~\cite{he2021dstc9} well-trained on clean written text is vulnerable to ASR errors. For instance, knowledge-seeking turn detection (KTD) F1 dropped dramatically from 0.9911 to 0.7602~\footnote{\url{https://github.com/alexa/alexa-with-dstc10-track2-dataset/blob/main/task2/baseline/README.md}}. Therefore, robust approaches are needed to bridge the mismatch.

In this paper, we address the robustness of knowledge selection with ASR errors by training a deep neural network to self-correct erroneous inputs and perform knowledge cluster classification in a joint fashion. First, we generate random errors from a proposed error simulator and inject the errors with corrective labels into data batches during model training. Second, we introduce knowledge cluster classification to quickly identify a subset of relevant documents from topically consistent knowledge clusters. Together with entity linking and tracking, document candidates are further reduced for re-ranking and deliver knowledge selection results.

Our contributions are three-fold: First, we propose an effective ASR error simulator learned from word confusion networks without human transcribed data. This implies that an ASR error simulator can be created and updated automatically by decoding speech audios. Second, we formulate knowledge selection with ASR errors using a joint model to self-correct simulated errors and perform knowledge cluster classification.
Third, we perform extensive experiments and show the effectiveness of our approach on the DSTC10 challenge.
\copyrightnotice

\section{Related Work}
\label{sec:related_work}
To overcome erroneous ASR utterances, many research have been focusing on ASR error detection~\cite{tam14_errordetect,yang19_errordetect} and error correction as individual modules, trying to detect and correct the errors so that the processed text are fed to downstream language understanding modules trained with clean text. \cite{namazifar21_interspeech} employed a warped language model, a variant of masked language model~\cite{devlin-etal-2019-bert}, to correct errors in ASR transcripts. \cite{wang2020asr_correction} proposed an augmented transformer to take in phonetic and orthographic information for entity correction using an error simulator based on the word N-gram confusion matrix. \cite{weng2020joint_asr_correction} leveraged aligned N-best lists as inputs to a deep model so that the model was optimized jointly for N-best re-ranking and language understanding. \cite{zhang2020spell_error_correct} proposed a soft-masked BERT to detect errors and mask out erroneous tokens with the mask-token embedding, followed by error correction in an end-to-end manner. These approaches have shown effectiveness but require high-quality human transcripts to be aligned with ASR transcripts to produce labels for model training. \cite{liu2020wcn_lu} flattened a word confusion network and used position embeddings and attention masks to encode the graph structure of a confusion network, trying to make use of word confusion information for robust language understanding.

Inspired by ASR error simulation~\cite{wang2020asr_correction} and masked language modeling~\cite{devlin-etal-2019-bert}, our approach addresses robust SLU in a joint modeling framework. Starting from clean dialogue, we employ an error simulator to randomly replace word tokens in an input utterance with noisy tokens during model training such that our model attempts to achieve two goals: 1) self-correct the noisy tokens; 2) use the ``repaired'' hidden states for classification. Unlike~\cite{wang2020asr_correction}, our error simulator does not require manual transcription for aligned labels. Instead, we analyze confusion networks and learn what words are confusing with each other from the ASR viewpoint. We leverage word confusion pairs to learn how a word is converted into another word via letter-to-letter alignment
(e.g ``deliver'' $\longrightarrow$ ``delver'' after replacing i by the deletion symbol *). Our error simulator does not ``hard-code'' the error patterns from the word confusion matrix but generates a variety of erroneous tokens during online data batching, enforcing a deep model to correct these ``negative'' tokens during training. In summary, we learn a Bert model so that simulated errors are self-corrected via a language modeling (LM) head, while the model simultaneously learns how to classify an input utterance. From the masked LM viewpoint~\cite{devlin-etal-2019-bert}, instead of masking tokens or replacing tokens with unrelated random tokens, we randomly replace input tokens in clean text with erroneous tokens that are ASR-error aware.

\section{Overall system architecture}
\label{sec:approach}
\begin{figure*}[htb]
  \centering
  \centerline{\includegraphics[scale=0.5]{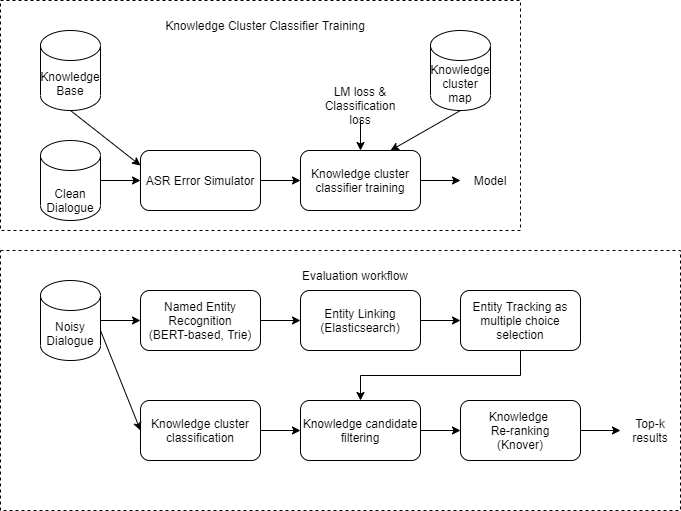}}
  \vspace{-2mm}
\caption{Proposed system workflow.}
\label{fig:overall_system}
\end{figure*}

The upper part of Figure~\ref{fig:overall_system} shows a workflow to train a robust knowledge cluster classifier. Our proposal relies on an ASR error simulator to generate a random error pattern followed by a joint model to correct errors and classify using LM and classification heads. Before knowledge selection, we introduce knowledge cluster classification to quickly identify relevant knowledge clusters.

The lower part of Figure~\ref{fig:overall_system} shows a system evaluation workflow. Given noisy dialogue, we first perform named entity recognition on each utterance turn using BERT-based NER model or Trie matching. Then we pass the detected knowledge named entities into entity linker based on elasticsearch where fuzzy search is supported. Then we approach the entity mention of the last turn of user as a multiple choice task over the detected linked knowledge named entities. In parallel, we classify noisy dialogue into knowledge clusters, so that the corresponding knowledge document candidates in the top predicted knowledge clusters are intersected with the top knowledge entity mentions during knowledge candidate filtering. Finally, the filtered knowledge candidates are re-ranked using Knover~\cite{he2021dstc9}.

\subsection{ASR error simulation}
\label{subsec:asr_sim}

Given speech audio, an ASR decoder decodes the audio and generates word lattices or N-best lists that are aligned into confusion networks~\cite{mangu99cn}. A confusion network provides alternative word candidates at each word position. For instance, a confusion network representation ``ok do they (delver $|$ deliver $|$ delo $|$ over $|$ lover $|$ del $|$ dolo)'' gives a word confusion set \{delver, deliver, delo, over, lover, del\} where the words sound confusing from an ASR decoder viewpoint. Here, we do not know which word token is correct in a confusion set. For creating an ASR error simulator, we focus on capturing and re-generating error patterns from confusion sets. In particular, we enumerate all possible word pairs from a confusion set, giving ``delver-deliver'', ``delver-delo'',...,``del-dolo''. We perform letter-to-letter alignment between each word pair so that position-dependent letter rewriting probabilities $Pr(t|s,i)$ are estimated based on co-occurrence statistics, where $i$ denotes the position of a source letter $s$ to be rewritten as letter $t$ including the deletion symbol ``*''. To generate a random erroneous word from a clean word:
\begin{enumerate}
    \item Randomly choose the number of edits uniformly from 1, 2 or 3.
    \item Randomly choose a letter position $i$ uniformly across the length of a word.
    \item Randomly choose an error type with $Pr(replacement)=0.9$ and $Pr(insertion)=0.1$.
    \item If the error type is replacement, draw a letter $t \sim Pr(t|s,i)$. Replace $s$ with $t$ in position $i$. 
    \item If the error type is insertion, draw a letter $t \sim Pr_*(t|s,i)$. Insert $t$ at position $i+1$.
    \item Repeat step 2 until the desired number of edits is achieved.
\end{enumerate}
Table~\ref{tbl:error_sim} shows samples of simulated errors from clean words. The simulated errors ``look'' similar to their clean words. The ability of simulating new errors is crucial to avoid overfitting a downstream SLU model to limited error patterns.


\begin{table}[htb]
\centering
\caption{Samples of ASR error simulation.}
\begin{tabular}{|c|c|}
\hline
clean word & Simulated versions \\ 
\hline
'alcohols' & 'alchols', 'alcohos', 'alcohls', 'alchos' \\
'alimentum' & 'alimentm', 'alimetum', 'alimenum' \\
'wifi' & 'wifr', 'wivi', 'wrfi', 'wifie' \\
\hline
\end{tabular}
\label{tbl:error_sim}
\end{table}

\subsection{Unsupervised knowledge clustering}
\label{subsec:kb_clustering}
There are 12k knowledge titles in the DSTC10 knowledge base. This poses a challenge of matching a dialogue against all knowledge titles, i.e. a brute foce approach, for practical knowledge selection. 
We observe that many entities in the knowledge base shared the same/similar titles. For example, ``A and B Guest House'' and ``Acorn Guest house'' share the same title regarding the ``wifi'' topic. This motivates us to design an iterative algorithm to group titles of similar topics into knowledge clusters in an unsupervised manner. Intuitively, titles with similar vocabulary (after removing stopwords and word normalization) formed initial knowledge clusters. For example, ``Does it have free wifi?'' is converted into ``free wifi'' after stopword removal and sorting. ``Is the wifi free'' is also converted into ``free wifi'' as well. Similarly, we also group similar bodies into clusters, assuming that a similar body (except the non-meaningful body) would imply a similar title, so that more knowledge clusters can be formed to enhance the variety of data for classifier training.

We train a BERT-based binary classifier to detect positive/negative title-body pairs. By default, all the title-body pairs in the knowledge base are considered as positive instances. In addition, similar titles in a knowledge cluster enrich the variety of positive instances by cross enumeration. We generate negative title-body pairs by randomly replacing a title with another title within the same entity, assuming that titles within an entity are mostly unique. Then we feed these positive and negative title-body instances to train a binary classifier.

Once a binary classifier is trained, our next step is to merge similar knowledge clusters.
We enumerate title-body pairs from each pair of knowledge clusters and classify the title-body pairs using our classifier. The classification results will give us an estimate about the similarity of the knowledge clusters. Pairs of knowledge clusters are combined when the majority of title-body pairs are positively classified according to a threshold. We repeat the grouping step iteratively until no more knowledge clusters can be formed. 
Additionally, we include a non-knowledge cluster containing user utterances that are non knowledge-seeking. In total, there were 95 knowledge clusters derived from 12k title-body pairs in the knowledge base covering hotel, restaurant, attraction, train and taxi domains. Table~\ref{tbl:kb_cluster} shows samples of a knowledge cluster for ``wifi'' and ``trail''.
 
\begin{table}[htb]
\centering
\caption{Samples of knowledge clusters.}
\begin{tabular}{|c|}
\hline
Are there any WiFi? \\
How much is the WiFi? \\
Do you offer free Wifi?\\
Does it offer free WiFi?\\
Do you have Wifi?\\
Hi! Does it offer any WiFi services?\\
\hline
What is the length of it? \\
How much time it will take to finish the guided tour?\\
How long is the trail?\\
How long does it take to complete a hike?\\
Is there any trail?\\
\hline
\end{tabular}
\label{tbl:kb_cluster}
\end{table}

\subsection{Knowledge cluster classification with self-correction}
\label{subsec:cid}
Inspired by masked LM, we randomly pick two words from a clean user utterance in a dialogue session, and apply error simulation in Section~\ref{subsec:asr_sim} to replace clean tokens with erroneous ones. We didn't replace more than 2 words in an utterance since injecting too much noise may corrupt the semantics of an utterance badly. With the generated errors and the corresponding clean word, we feed an erroneous dialogue utterance into a deep model, e.g. Bert~\cite{devlin-etal-2019-bert} and learn to detect and correct the errors using the LM head.

\begin{figure}[htb]
  \centering
  \centerline{\includegraphics[scale=0.4]{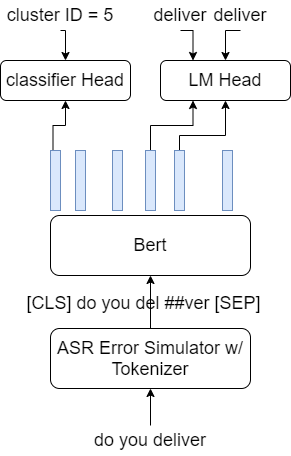}}
  \vspace{-2mm}
\caption{Proposed classifier training with an ASR error simulator.}
\label{fig:proposed}
\end{figure}

Using multi-task learning, our ultimate target is to train a robust SLU model to detect and correct errors, and to classify an utterance with a knowledge cluster label. Given joint model parameters $\Theta=\{W_{bert}, W_{lm}, W_{cls}\}$ where $\Theta$ denotes parameters of a pre-trained Bert model, the language modeling head and the classification head respectively, our joint loss function becomes:
\begin{equation}
L(\Theta)
= L_{cls}(W_{bert}, W_{cls}) + \lambda \cdot L_{lm}(W_{bert}, W_{lm})
\end{equation}
where $\lambda$ denotes the LM weight for correcting the ASR errors. When $\lambda$ is set to zero, the model just trains with noise, ignoring the error-correction mechanism. Figure~\ref{fig:proposed} shows the proposed system architecture with double heads.

%
%

\section{Experiments}
\label{sec:expt}

\subsection{Data}
\label{subsec:data}
We employed datasets from the Dialogue System Technology Challenge, namely DSTC9~\cite{kimdstc9} and DSTC10~\cite{kimdstc10}. The challenge had 3 sub-tasks: 1) Knowledge-seeking turn detection (KTD) to determine if the last user turn had knowledge-seeking intention. For example, after a system returned a list of hotel recommendations that satisfy user's requirements, users may ask questions like: ``does this hotel provide free wifi?'' as positive knowledge seeking; 2) Knowledge-selection (KS) to search for a document from an external knowledge base to answer the question. In multi-turn dialogues, entity linking and tracking was required during searching for relevant documents.
In this paper, we mainly reported results for KTD and KS. DSTC9 datasets contained clean dialogue while the DSTC10 dataset had noisy dialogue with ASR errors.
According to the track organizer, their ASR decoder in DSTC10 was based on wav2vec 2.0~\cite{wav2vec2020baevski} pre-trained on 960 hours of Librispeech and fine-tuned with 10\% of in-domain speech data with word error rate of 24.09\%. DSTC10 dataset came with N-best lists. We used SRILM toolkit~\cite{Stolcke2002SRILMA} to generate confusion networks from N-best lists. Since only user utterances contained ASR errors while system utterances were clean, we simulated errors on user utterances only. Last-turn user utterances were considered for error simulation and model training. Table~\ref{tbl:asr_errors} shows samples of knowledge-seeking user utterances containing ASR errors and their clean versions.

\begin{table}[htb]
\centering
\caption{Samples of erroneous user utterances.}
\begin{tabular}{|c|}
\hline
perfect and what is the check {\bf kin} time for that\\
ok do they {\bf delver} food\\
o ea do you know if that's a good place uh for \\ 
{\bf marge} groups\\
uh {\bf taff ally} uh do they offer daily house keeping\\
\hline
perfect and what is the check in time for that\\
do they deliver food\\
by chance do you know if that's a good place uhhh for \\
large groups\\
uhhh possibly uhhh do they offer daily house keeping\\
\hline
\end{tabular}
\label{tbl:asr_errors}
\end{table}

One difficulty of our work was that the provided DSTC10 dataset had only 263 dialogue sessions. For model tuning and evaluation, we split them into two equal parts. The first half was used as a development set while the second half was used as a test set.

To create our training set, we employed the DSTC9 evaluation set comprising only clean dialogue sessions. We didn't use the DSTC9 training and validation sets for our model training because they did not give any improvement. Since the DSTC10 dataset had overlapped dialogues with the DSTC9 evaluation set, we excluded the overlapped dialogues from training, yielding 3867 training dialogues. To enhance data coverage, we augmented titles from the knowledge base as the last ``user-turn'' into our training set, yielding 15906 training dialogues as our final training set.

\subsection{Model training}
The motivation of using the knowledge clusters as labels was to avoid brute-force evaluation of the relevance between a last-turn user utterance and possible documents in the knowledge base.
To ensure good quality of knowledge clusters, we manually inspected and adjusted the knowledge clusters so that questions within each cluster were topically consistent.
Then, we addressed knowledge selection in two steps. First, we predicted the knowledge cluster of the last user turn using the joint Bert classifier trained with error simulation and correction in Section~\ref{subsec:cid}. This step narrowed down the document candidates based on top-K predicted clusters.
To pinpoint relevant documents further, we built entity-linking and entity-tracking systems based on techniques such as Trie-based and Bert-based named entity recognition, fuzzy match in Elasticsearch, and multiple-choice-headed~\cite{Radford2018ImprovingLU} Bert for entity tracking. 

We trained Bert (bert-base-uncased) without ASR error simulation as our baseline. Then we trained joint models with different LM weights ranging from 0 (trained with noise without error correction), 0.1, 0.5, and 1.0. All models shared the same modeling architecture and were trained using the same hyper-parameters. We used the Huggingface Transformers library~\cite{wolf-etal-2020-transformers} for pretrained models and the DSTC9 code base~\footnote{\url{https://github.com/alexa/alexa-with-dstc9-track1-dataset}} for our implementation.
We trained all models with 100 epochs with batch size of 16 on a single GPU. We used Adam~\cite{kingma2014adam} for optimization with a learning rate of 6.25e-5 and  $\epsilon$ of 1e-8. Gradient clipping was applied with a maximum gradient norm of 1.0. Linear scheduling without warm-up was used. All random seeds were set to 42. For knowledge cluster classification, we only used the last user utterance since performance was hurt using all dialogue turns.
The last user turn usually carried the most crucial information for knowledge cluster classification.

\subsection{Results}
\label{sec:ks_result}
Table~\ref{tbl:cid_dev} shows the knowledge cluster classification results using the proposed approach with various LM weights on the development set. First, the proposed approach outperformed the baseline on all metrics. In particular, Recall@1 was boosted drastically from 0.8118 to 0.94 when the LM weights were set to 0.1 or 0.5. Tuning the LM weight was crucial as performance dropped LM weights were set to 0.0 and 1.0. It was surprising that simply adding noise without self-correction still improved performance. On the other hand, we observed that its model training would result in early stopping based on the optimal performance on the development set.
In contrast, models took more steps to train when error correction was considered. This may be because the models took more effort to minimize the joint loss function with more variety of simulated errors. 
As a by-product of knowledge cluster classification, the proposed model addressed knowledge-seeking turn detection (KTD) when the predicted label was either zero or non-zero. Consistently, KTD F1 was also improved from 0.9306 to 0.96 compared to the baseline.


\begin{table}[H]
\centering
\caption{Knowledge-turn detection and knowledge cluster classification results on the development set using the proposed approach with various LM weights. MMR: Mean Reciprocal Rank.}
\tabcolsep=0.11cm
\begin{tabular}{|c|c|c|c|c|c|}
\hline
LM weight     & F1 (KTD)     & R@1     & R@5        & MRR@5  & Steps \\
\hline
Baseline      & 0.9306      & 0.8118   & 0.8712     & 0.8333 & 14925 \\
\hline
0.0           & {\bf 0.96}      & 0.90   & {\bf 0.94}     & 0.915  & 13930 \\
0.1           & {\bf 0.96}      & {\bf 0.94}   & {\bf 0.94}     & {\bf 0.94}   & 84575 \\ 
0.5           & {\bf 0.96}      & {\bf 0.94}   & {\bf 0.94}     & {\bf 0.94}   & 59700 \\ 
1.0           & 0.94      & 0.92   & 0.92     & 0.92   & 32835 \\ 
\hline
\end{tabular}
\label{tbl:cid_dev}
\end{table}

Table~\ref{tbl:cid_test} shows the knowledge cluster classification results on the test set. We observed consistent improvement on all metrics. In particular, we observed a drastic improvement in Recall@1 from 0.7924 to 0.9333 when the optimal LM weight according to the development set. KTD F1 was also boosted significantly from 0.9433 to 0.9904. Performance trends on the development and test sets were similar, demonstrating the effectiveness of the proposed approach on noisy utterances with ASR errors.

\begin{table}[H]
\centering
\caption{Knowledge-turn detection and knowledge cluster classification results on the test set using the proposed approach with various LM weights.}
\tabcolsep=0.11cm
\begin{tabular}{|c|c|c|c|c|}
\hline
LM weight              & F1 (KTD)     & R@1          & R@5 & MRR@5  \\
\hline
Baseline            & 0.9433       & 0.7924       & 0.8301       & 0.8056        \\
\hline
0.0   & 0.9714  & 0.8380 & 0.8571 & 0.8476  \\ 
0.1   & 0.9714  & 0.8952 & 0.8952 & 0.8952  \\
0.5   & {\bf 0.9904}  & {\bf 0.9333} & {\bf 0.9333} & {\bf 0.9333}  \\ 
1.0   & 0.9411  & 0.9019 & 0.9019 & 0.9019  \\ 
\hline
\end{tabular}
\label{tbl:cid_test}
\end{table}

\begin{table}[H]
\centering
\caption{Knowledge-turn detection and knowledge selection results on the test set using the proposed approach with various LM weights after knowledge re-ranking using Knover~\cite{he2021dstc9}}
\tabcolsep=0.11cm
\begin{tabular}{|c|c|c|c|c|}
\hline
LM weight           & F1 (KTD)& R@1    & R@5    & MRR@5  \\
\hline
Baseline           & 0.9433  & 0.7358 & 0.8301 & 0.7798  \\
\hline
0.0  & 0.9714  & 0.7238 & 0.8571 & 0.7873  \\ 
0.1  & 0.9714  & 0.7619 & 0.8952 & 0.8253  \\
0.5  & {\bf 0.9904}  & {\bf 0.7809} & {\bf 0.9333} & {\bf 0.8460}  \\ 
1.0  & 0.9411  & 0.7647 & 0.9019 & 0.8300  \\
\hline
0.5 (w/o entity tracker) & {\bf 0.9904}  & 0.6666 & 0.8380 & 0.7333  \\
0.5 (w/o Knover)   & {\bf 0.9904}  & 0.7619 & 0.8952 & 0.8101  \\
\hline
\end{tabular}
\label{tbl:ks_test}
\end{table}

Finally, we combined the top-K knowledge cluster predictions with our multiple-choice-headed~\cite{Radford2018ImprovingLU} Bert-based entity tracker, yielding 
pre-selected knowledge document candidates for further re-ranking. 
Re-ranking was crucial because knowledge clusters were coarse and may contain documents from different sub-topics 
as shown in Table~\ref{tbl:kb_cluster}. 
For each system, we fed their pre-selected knowledge document candidates into the DSTC9 winner system (Knover) for knowledge re-ranking~\cite{he2021dstc9}.

Table~\ref{tbl:cid_dstc10eval} shows knowledge cluster classification results on our proposed systems. For comparison, we took the Knover predicted outputs from DSTC10 and mapped the document ID to knowledge cluster ID. Our submitted system outperformed Knover on all metrics. Table~\ref{tbl:ks_test} shows knowledge turn detection and knowledge selection results on the test set. Consistently, the proposed approach showed improved performance across all metrics. In particular, our approach yielded 4.5\%, 10\% and 6.62\% absolute improvement on Recall@1, Recall@5 and MRR@5 respectively compared to the baseline. We also compared the case where our entity-tracker was disabled so that the Knover system performed knowledge re-ranking directly from documents in the predicted top-K knowledge clusters. However, the results were worse as shown in Table~\ref{tbl:ks_test}, showing that our entity-linking and entity tracker were effective in removing a lot of irrelevant entities from re-ranking. In a separate experiment, re-ranking with Knover was still effective compared to the un-ranked documents within the tracked top entities sorted by document IDs. 


\begin{table}[H]
\centering
\vspace{-3mm}
\caption{DSTC10 evaluation results on knowledge cluster classification using the proposed approach.}
\begin{tabular}{|c|c|c|c|}
\hline
System             &  R@1    & R@5    & MRR@5  \\
\hline
Knover baseline & 0.6923  & 0.7374 & 0.7091  \\
\hline
Submitted system &  {\bf 0.7901} & {\bf 0.8263} & {\bf 0.8045}  \\ 
\hline
Post-eval (last-turn) & 0.7838 & 0.8256 & 0.7998  \\ 
Post-eval (all-turns) & 0.6973 & 0.7496 & 0.7170  \\ 
\hline
\end{tabular}
\label{tbl:cid_dstc10eval}
\end{table}

\begin{table}[H]
\centering
\vspace{-3mm}
\caption{DSTC10 evaluation results on knowledge-turn detection using the proposed approach.}
\begin{tabular}{|c|c|c|c|}
\hline
System             & P  & R  & F1  \\
\hline
Knover baseline & 0.8967 & 0.6735 & 0.7692  \\
\hline
Submitted system & 0.8852 & 0.8697 & 0.8774  \\ 
\hline
Post-eval (last-turn) & {\bf 0.8878} & 0.8696 & {\bf 0.8786}  \\ 
Post-eval (all-turns) & 0.7636 & {\bf 0.9033} & 0.8276    \\ 
Unsup knowledge clusters & 0.8494 & 0.9004 & 0.8742   \\ 
\hline
\end{tabular}
\label{tbl:ktd_dstc10eval}
\end{table}

\begin{table}[H]
\centering
\vspace{-3mm}
\caption{DSTC10 evaluation results on knowledge selection using the proposed approach.}
\begin{tabular}{|c|c|c|c|c|c|c|}
\hline
System             &  R@1    & R@5    & MRR@5  \\
\hline
Knover baseline &  0.495  & 0.6472 & 0.5574  \\
\hline
Submitted system &  0.7105 & 0.7976 & 0.7493  \\ 
\hline
Post-eval (last-turn) &  {\bf 0.7144} & {\bf 0.8032} & {\bf 0.7541}  \\ 
Post-eval (all-turns) &  0.6586 & 0.7391 & 0.6950  \\ 
Unsup knowledge clusters &  0.6979 & 0.7846 & 0.7369  \\ 
\hline
\end{tabular}
\label{tbl:ks_dstc10eval}
\end{table}

Table~\ref{tbl:ktd_dstc10eval} and Table~\ref{tbl:ks_dstc10eval} show the DSTC10 evaluation results with 1988 test dialogues on knowledge turn detection and knowledge selection respectively. Our proposed approach shows effectiveness in all metrics using a single model. In particular, Recall@1 was boosted dramatically from 0.495 to 0.7105 compared to the Knover baseline~\cite{he2021dstc9} trained only on clean DSTC9 dialogues.
Among the 16 teams with their best submission, our best system ranked the 3rd place in Recall@1 and Recall@5 for knowledge selection sub-task. Due to time limits, our submitted system was a single model without model ensembles.

After evaluation, we fixed some errors in our system workflow and reran the training and evaluation. We obtained further improvement on all metrics, boosting the Recal@1 from 0.7105 to 0.7144. In addition, we used all dialogue turns to train the knowledge cluster classifier. However, we found a significant degradation in all metrics, showing that including the user and system turns hurt performance, especially when the user turns contain speech recognition errors. We also re-trained the knowledge cluster classifier without human adjustment of the knowledge cluster labels. Within our expectation, we observed some degradation since the knowledge clusters from unsupervised clustering contained some errors.

\section{Discussion}
\label{sec:discuss}
For ASR error simulation, at first we replaced a clean word randomly by sampling a noisy word from the word confusion matrix. However, knowledge cluster classification results on the test set were much worse with Recall@1 only at 0.8118 compared to 0.9333. This may be due to the lack of ASR error coverage in the word confusion matrix to cope with unseen error patterns in tests.

Having a trainable ASR error simulator was attractive. We tried finetuning GPT2~\cite{Radford2018ImprovingLU} with word confusion pairs and generating words letter by letter using an LM head or a pointer generator head~\cite{see2017ptrgen, enarvi-etal-2020-generating}. However, none of these approaches could generate reasonable variants of new words. The lack of training data may be the cause. With more ASR decoded training data, we would further investigate it in the future.

The proposed error simulator is limited to editing a clean word to generate errors. Moreover, it is hard for our simulator to generate phonetically similar words or phrases such as ``kommen toor'' from ``coit tower''. It may be helpful to integrate a phonetic dictionary~\cite{wang2020asr_correction} to improve the variety of the simulated errors.

Lastly, it would be interesting to evaluate our work on dialogue state tracking with ASR errors. In addition, training a knowledge selection ranker with a proposed ASR simulator will be our future work as well.

\section{Conclusions}
We have proposed ASR error simulation to improve knowledge selection in knowledge-grounded task oriented dialogues. Our approach does not require human transcription but leverage word confusion networks to build an error simulator and simulate a variety of errors for robust model training. We have proposed a joint modeling approach using double heads to correct the simulated errors and perform knowledge cluster classification. Our proposed approach has shown significant improvement in knowledge-turn detection and knowledge selection on our test set and the DSTC10 official evaluation set.

\vfill\pagebreak

\bibliography{strings,refs}

\end{document}